\pdfoutput=1
\documentclass{article}

\usepackage{spconf}

\usepackage{times}
\usepackage{latexsym}
\usepackage[T1]{fontenc}
\usepackage[utf8]{inputenc}
\usepackage{microtype} 
\usepackage{inconsolata}
\usepackage{graphicx}
\usepackage{amsmath,amssymb}
\usepackage{hyperref}
\usepackage{cleveref}
\usepackage{enumitem}

\usepackage{adjustbox}
\usepackage{xspace}
\usepackage{xstring}

\newcommand{\ours}{\textbf{\textsc{ReasonAudio}}\xspace}
\newcommand{\textaudio}{Text--Audio\xspace}
\newcommand{\eg}{\hbox{\emph{e.g.,}}\xspace}

\usepackage[caption=false]{subfig}

\usepackage{tabularx}
\usepackage{booktabs}
\usepackage{multirow}
\usepackage{colortbl}
\usepackage{caption}
\title{REASONAUDIO: A BENCHMARK FOR EVALUATING REASONING BEYOND MATCHING\\IN TEXT--AUDIO RETRIEVAL}

\name{Honglei Zhang$^{1}$, Yuting Chen$^{2}$, Chenpeng Hu$^{2}$,
Pengfei Zhou$^{2}$, Siyue Zhang$^{3,*}$, Yilei Shi$^{2,*}$
\thanks{\llap{\textsuperscript{*}\,}Corresponding authors.}}

\address{
$^{1}$School of Software, Nanjing University, China\\
$^{2}$School of Software, Northwestern Polytechnical University, China\\
$^{3}$College of Computing and Data Science,
Nanyang Technological University, Singapore
}

\begin{document}
\ninept
\maketitle
\begin{abstract}

Existing audio retrieval benchmarks primarily assess semantic matching, lacking comprehensive evaluation of the logical reasoning capabilities required by complex queries. We introduce \ours\footnote{Project: \url{https://huggingface.co/datasets/ReasonAudio}}, a benchmark for reasoning-intensive \textaudio Retrieval that evaluates four abilities: negation, temporal order, sound co-occurrence, and sound duration. It comprises five synthetic subtasks with 1,000 queries over 10,000 composite audio clips and a natural subtask with 100 queries over 1,000 real-world clips. Evaluation of 11 state-of-the-art retrieval systems reveals substantial limitations: the best-performing model, OmniEmbed-7B, achieves an overall score of 20.7. In a controlled experiment designed to reduce the influence of sound-event matching, OmniEmbed-7B attains 53.8\% average accuracy, compared with 70.6\% for its generative backbone, Qwen2.5-Omni-7B-Thinker, and 95.6\% for humans. Our results highlight the challenges of reasoning-intensive audio retrieval and reveal a performance gap between OmniEmbed-7B and its generative backbone.

\end{abstract}




\section{Introduction}

The rapid proliferation of multimodal content has made multimodal retrieval increasingly important for modern multimedia systems. Among its various tasks, \textaudio Retrieval aims to locate relevant audio clips from large-scale corpora based on textual queries, supporting applications such as media search, content organization, and intelligent assistants. 

Existing \textaudio retrieval methods can be broadly categorized into three paradigms. First, two-stage pipelines convert audio into text using captioning models~\cite{huang2025stepaudiounifiedunderstandinggeneration,chu2024qwen2audiotechnicalreport} and then perform retrieval with text embedding models~\cite{chen2025m3embeddingmultilingualitymultifunctionalitymultigranularity,zhang2025qwen3embeddingadvancingtext}. Second, direct embedding methods adopt CLIP-style contrastive learning~\cite{clip} to align audio and text in a shared embedding space~\cite{elizalde2022claplearningaudioconcepts,guzhov2021audioclipextendingclipimage,wu2022wav2cliplearningrobustaudio}. Third, recent multimodal embedding approaches leverage pretrained MLLMs to learn unified cross-modal representations for retrieval~\cite{xiao2025scalinglanguagecentricomnimodalrepresentation,e5omni}.



Despite these advances, most existing \textaudio Retrieval benchmarks emphasize semantic matching (Table~\ref{tab:audio_benchmarks}), with limited coverage of the reasoning abilities required by complex real-world queries, such as temporal reasoning and negation understanding~\cite{bright,mrag}. Such reasoning abilities have received increasing attention in audio-language models, with benchmarks such as CompA~\cite{ghosh2024compa} and PolyBench~\cite{poly} assessing compositional understanding. However, the evaluation of these abilities remains limited in retrieval settings. In Text--Audio Retrieval, prior work introduced SynCaps~\cite{temporal-audio} to evaluate temporal order understanding. To provide a more comprehensive evaluation of reasoning abilities in retrieval, we introduce \ours, a reasoning-intensive \textaudio Retrieval benchmark covering four core abilities: negation, temporal order, sound co-occurrence (overlap), and sound duration. It comprises five synthetic subtasks and one natural subtask. As shown in Figure~\ref{fig:fig1}, we synthesize 10,000 temporally controlled composite clips from 200 atomic sounds and generate 1,000 template-based queries. Relevance labels are derived from composition metadata, with additional GPT-5.5 consistency checks and human listening evaluation. The natural subtask includes 100 manually written queries over 1{,}000 real-world clips from AudioSet-Strong~\cite{hershey2021temporallystrong}.

\begin{figure}[t!]
\centering
\includegraphics[width=\columnwidth]{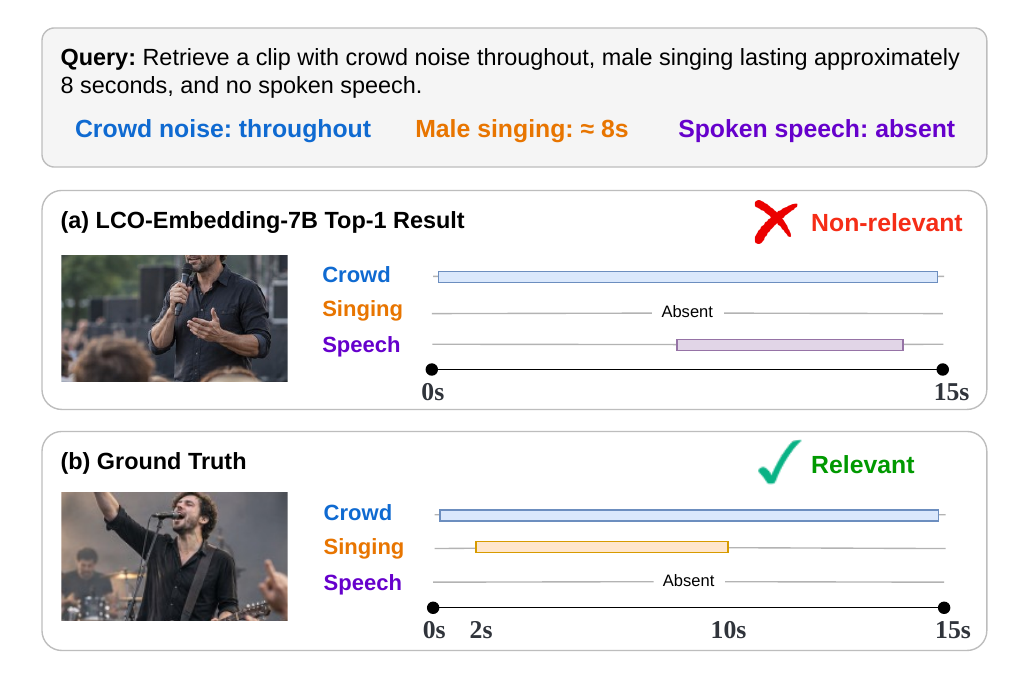}
\caption{A failure case illustrating how an MLLM-based embedding model struggles to interpret negation and sound duration.}
\label{fig0}
\end{figure}

\begin{figure*}[t!]
 \centering
 \includegraphics[width=\textwidth]{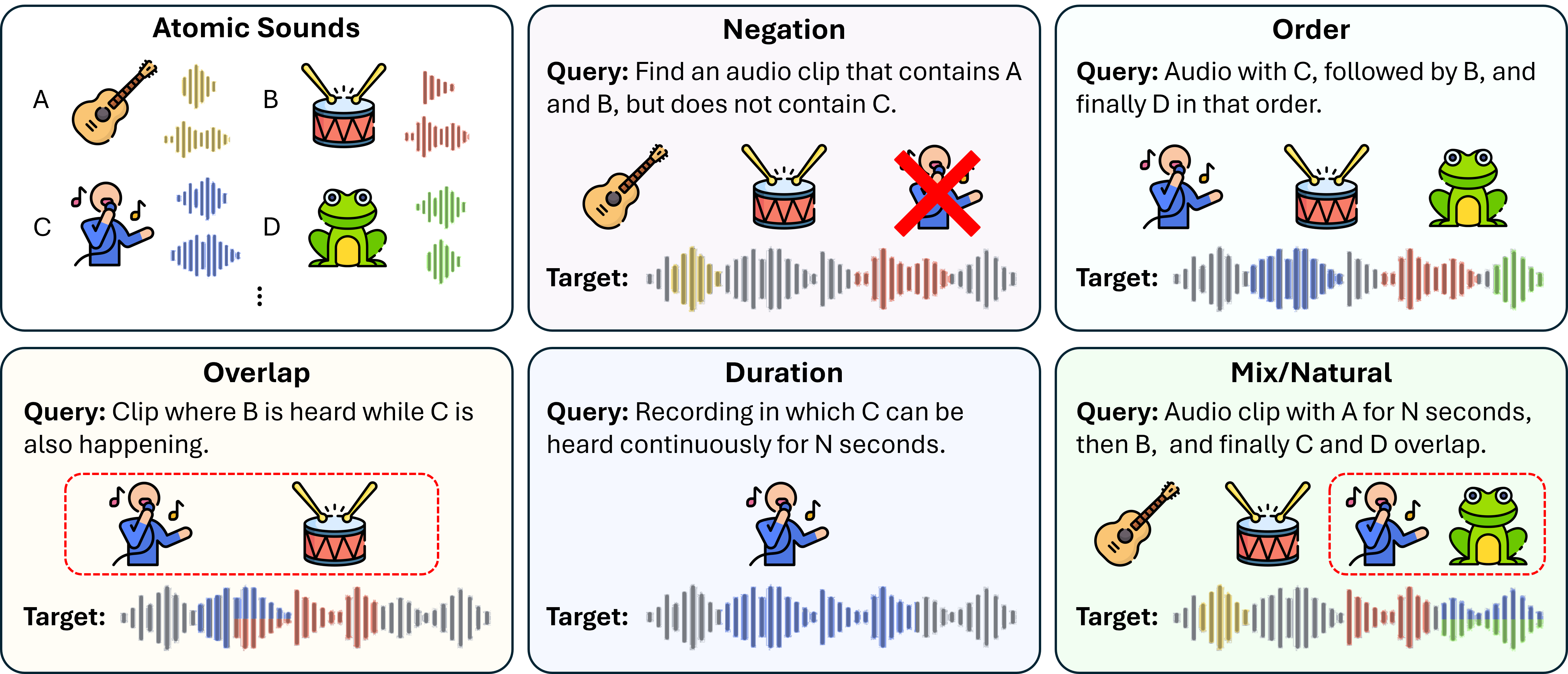}
 \caption{Illustration of the \ours benchmark. The first panel shows atomic sounds, and the remaining five panels illustrate the six subtasks, with Mix and Natural sharing the final panel to demonstrate their common combined-constraint query structure.}
 \label{fig:fig1}
\end{figure*}

We conduct a comprehensive evaluation of 11 state-of-the-art retrieval systems on \ours, spanning two-stage \textaudio retrieval approaches, CLIP-style embedding models, and MLLM-based embedding models. Reasoning-intensive audio retrieval remains challenging for all evaluated systems: the best-performing model, OmniEmbed-7B, achieves an overall score of only 20.7. All evaluated models struggle with negation and duration constraints, as exemplified in Figure~\ref{fig0}. To reduce the influence of sound-event matching, we conduct a controlled two-choice experiment in which both candidates contain the required sound events, but only one satisfies the specified logical constraints. OmniEmbed-7B achieves 53.8\% accuracy, compared with 70.6\% for its generative backbone, Qwen2.5-Omni-7B-Thinker. This gap suggests that strong performance in generative reasoning does not necessarily translate into equally strong performance in embedding-based retrieval. We hope that \ours and these insights will guide future advances in \textaudio Retrieval.





\begin{table}[b!]
\centering
\caption{Comparison of \ours with existing datasets designed for or adaptable to audio-related retrieval tasks. The final column reports the number of evaluation queries.}
\label{tab:audio_benchmarks}
\renewcommand{\arraystretch}{1.05}
\setlength{\tabcolsep}{3pt}
\scriptsize
\begin{adjustbox}{max width=\columnwidth}
\begin{tabular}{@{}lccr@{}}
\toprule
\textbf{Benchmark} & \textbf{Modality} & \textbf{Retrieval Task} & \textbf{\#Qry.} \\
\midrule
AudioCaps~\cite{audiocaps} & Text $\leftrightarrow$ Audio & Audio caption retrieval & 4.9K \\
Clotho~\cite{drossos2019clothoaudiocaptioningdataset} & Text $\leftrightarrow$ Audio & Audio caption retrieval & 5K \\
SoundDescs~\cite{Koepke_2023} & Text $\leftrightarrow$ Audio & Audio caption retrieval & 4.9K \\
AudioSet-Strong~\cite{hershey2021temporallystrong} & Audio $\rightarrow$ Text & Sound event detection & 14K \\
SVQ (MSEB)~\cite{heigold2026massivesoundembeddingbenchmark} & Audio $\rightarrow$ Text & Voice search & 181K \\
Spoken SQuAD~\cite{li2018spokensquadstudymitigating} & Text $\rightarrow$ Audio & Spoken document retrieval & 5.3K \\
SynCaps~\cite{temporal-audio}
& Text $\rightarrow$ Audio
& Temporal order-focused retrieval
& 0.5K \\
\midrule
\textbf{\ours} & Text $\rightarrow$ Audio & Negation, order, overlap, duration & 1.1K \\
\bottomrule
\end{tabular}
\end{adjustbox}
\end{table}

\section{Related Work}

\subsection{\textaudio Retrieval}
\textaudio Retrieval aims to identify relevant audio clips given a natural-language query. Common benchmarks built from AudioCaps and Clotho~\cite{audiocaps,drossos2019clothoaudiocaptioningdataset} primarily evaluate semantic matching between audio and text. Beyond semantic matching, CompA~\cite{ghosh2024compa} and PolyBench~\cite{poly} have investigated compositional reasoning capabilities for audio-language models, whereas our focus is retrieving relevant clips from a large audio corpus. Temporal ordering has also been studied in \textaudio Retrieval~\cite{temporal-audio}, motivating broader evaluation across multiple logical and temporal constraints. Existing retrieval approaches include audio captioning followed by text retrieval~\cite{huang2025stepaudiounifiedunderstandinggeneration,chu2024qwen2audiotechnicalreport,chen2025m3embeddingmultilingualitymultifunctionalitymultigranularity}, contrastive audio--text representation learning~\cite{elizalde2022claplearningaudioconcepts,guzhov2021audioclipextendingclipimage,wu2022wav2cliplearningrobustaudio}, and embeddings adapted from MLLMs~\cite{xiao2025scalinglanguagecentricomnimodalrepresentation,e5omni}. Complementing prior compositional and temporal evaluations, \ours benchmarks all three retrieval paradigms across negation, order, overlap, duration, and combined constraints using both controlled synthetic and natural audio.


\subsection{Reasoning-Intensive Retrieval}

Reasoning-intensive retrieval has attracted increasing attention, focusing on capabilities beyond lexical or semantic similarity, such as temporal reasoning, negation understanding, and instruction following~\cite{bright,temporal-audio,mrag}. Prior studies show that even recent embedding models struggle when relevance depends on satisfying multiple logical constraints. However, this challenge remains underexplored in \textaudio retrieval, where systems must recognize acoustic events and reason about their absence, order, overlap, and duration. \ours complements prior work by extending reasoning-intensive retrieval to the audio domain and evaluating these abilities using controlled synthetic clips and natural real-world recordings.


\section{Task Formulation}
We formally define the \textaudio Retrieval task as follows. Let $Q = \{q_1,\dots,q_n\}$ be text queries and $D = \{d_1,\dots,d_m\}$ the audio corpus. For query $q_i \in Q$, each candidate $d_j \in D$ is either relevant ($d_+$) or non-relevant ($d_-$). The goal is to rank relevant audio clips $d_+$ above irrelevant clips $d_-$ for each query $q_i$ using a scoring function $f(q_i,d_j)$. \ours evaluates four core reasoning abilities---negation, order, overlap, and duration---through the following six subtasks. Figure~\ref{fig:fig1} illustrates the query structures of the six subtasks, with a shared schematic for Mix and Natural.

\begin{table*}[t!]
\centering
\caption{Comparison of SOTA \textaudio Retrieval models across the six \ours subtasks. Bold and underlined values indicate the best and second-best model results, respectively. \textsuperscript{*}Human evaluation assesses subtask solvability using OmniEmbed-7B’s top-6 candidates, replacing one with a ground-truth positive when none is present. Its scores are not directly comparable to full-corpus retrieval results.}
\label{tab:main_results}
\renewcommand{\arraystretch}{1}

\begin{adjustbox}{max width=\textwidth}
\setlength{\tabcolsep}{2.5pt}
\begin{tabular}{@{}lccccccccccccc@{}}
\toprule
\multirow{2}{*}{\textbf{Model}}
& \multicolumn{2}{c}{\textbf{Negation}}
& \multicolumn{2}{c}{\textbf{Order}}
& \multicolumn{2}{c}{\textbf{Overlap}}
& \multicolumn{2}{c}{\textbf{Duration}}
& \multicolumn{2}{c}{\textbf{Mix}}
& \multicolumn{2}{c}{\textbf{Natural}}
& \textbf{Overall} \\
\cmidrule(lr){2-3} \cmidrule(lr){4-5} \cmidrule(lr){6-7}
\cmidrule(lr){8-9} \cmidrule(lr){10-11} \cmidrule(lr){12-13}
& \small Acc@1 & \small nDCG@10
& \small Acc@1 & \small nDCG@10
& \small Acc@1 & \small nDCG@10
& \small Acc@1 & \small nDCG@10
& \small Acc@1 & \small nDCG@10
& \small Acc@1 & \small nDCG@10
& \\
\midrule

\multicolumn{14}{c}{\textit{Two-Stage \textaudio Retrieval}} \\
\midrule
Qwen2-Audio+BGE-M3      & 0.0 & 0.0 & 0.5 & 1.7 & 0.0 & 0.0 & 0.5 & 0.5 & 0.0 & 0.0 & 8.0 & 11.0 & 1.9 \\
Qwen2-Audio+Qwen3-Emb.  & 0.0 & 0.0 & 0.5 & 0.5 & 0.0 & 0.0 & 1.5 & 0.7 & 0.0 & 0.0 & 8.0 & 13.0 & 2.0 \\
Step-Audio+BGE-M3       & 2.0 & 1.4 & 0.5 & 0.5 & 2.5 & 3.8 & 8.5 & 7.4 & 1.0 & 2.4 & 2.0 & 1.6 & 2.8 \\
Step-Audio+Qwen3-Emb.   & 4.0 & 2.7 & 1.5 & 2.4 & 4.0 & 5.8 & 6.0 & 7.5 & 3.0 & 4.9 & 1.0 & 0.4 & 3.6 \\
\midrule

\multicolumn{14}{c}{\textit{CLIP-style Embedding Models}} \\
\midrule
CLAP      & 2.0 & 1.2 & 3.5 & 6.6 & 9.5 & 17.8 & \underline{11.0} & 9.5 & 6.0 & 12.5 & \textbf{19.0} & \underline{30.2} & 10.7 \\
AudioCLIP & 0.5 & 0.1 & 1.5 & 0.6 & 1.0 & 4.0 &
2.5 & 3.0 & 0.0 & 0.0 & 0.0 & 0.0 & 1.1 \\
Wav2CLIP  & 0.0 & 0.2 & 0.0 & 0.0 & 0.0 & 0.0  & 1.0 & 1.8 & 0.0 & 0.2  & 1.0 & 0.9 & 0.4 \\
\midrule

\multicolumn{14}{c}{\textit{MLLM-based Embedding Models}} \\
\midrule
LCO-Embedding-3B & 3.0  & 5.0  & 16.0 & 26.2 & \underline{21.5} & \underline{29.5} & 10.5 & \underline{10.5} & 6.0  & 14.8 & 16.0 & 28.8 & 15.7 \\
LCO-Embedding-7B & 2.0  & 2.8  & \underline{23.0} & \underline{31.8} & \textbf{29.5} & \textbf{40.0} & \textbf{14.5} & \textbf{10.6} & \underline{15.5} & 25.1 & 10.0 & 17.5 & \underline{18.5} \\
e5-omni-7B       & \textbf{10.0} & \textbf{10.0} & 22.0 & 30.2 & 7.0  & 12.3 & 9.0  & 8.4  & 15.0 & \underline{25.7} & \underline{18.0} & 29.9 & 16.5 \\
OmniEmbed-7B     & \underline{5.5}  & \underline{7.0}  & \textbf{24.5} & \textbf{36.2} & 18.0 & 26.4 & 6.5  & 8.5  & \textbf{28.0} & \textbf{40.1} & 16.0 & \textbf{31.5} & \textbf{20.7} \\
\midrule
Human Evaluation\textsuperscript{*}
& 100.0 & \textemdash
& 82.0  & \textemdash
& 94.0  & \textemdash
& 100.0 & \textemdash
& 94.0  & \textemdash
& 90.0  & \textemdash
& \textemdash \\
\bottomrule
\end{tabular}
\end{adjustbox}
\end{table*}

\begin{itemize}
    \setlength{\topsep}{2pt}
    \setlength{\partopsep}{0pt}
    \setlength{\itemsep}{4pt}
    \setlength{\parsep}{0pt}
    \setlength{\parskip}{0pt}

    \item \textbf{Negation:} Retrieve clips containing target sound events while excluding specified events.

    \item \textbf{Order:} Retrieve clips in which sound events occur in a specified temporal order.

    \item \textbf{Overlap:} Retrieve clips in which specified sound events overlap temporally.

    \item \textbf{Duration:} Retrieve clips containing sound events of a specified duration.

    \item \textbf{Mix/Natural:} Retrieve clips satisfying combinations of the above constraints.
\end{itemize}









\section{REASONAUDIO BENCHMARK}

To evaluate reasoning beyond semantic matching in \textaudio retrieval, we introduce \ours, a benchmark comprising six subtasks with 1,100 queries and a corpus of 11,000 audio clips. This section details the dataset construction pipeline.

\subsection{Synthetic Subtask Construction}

To construct the audio corpus for the synthetic subtasks, we curated 200 atomic sound samples from publicly licensed recordings in FSD50K~\cite{fonseca2022fsd50kopendatasethumanlabeled} and Freesound~\cite{jiang2025freeaudiotrainingfreetimingplanning}. Each sample contains a single, clearly identifiable sound event. We selected high-quality clips and manually verified their event labels by listening to each recording, ensuring that each event label refers to a single, unambiguous sound category. We then standardized the annotations and filenames, extracted clean, continuous segments containing the target event, and converted all clips to mono WAV format. To reduce ambiguity in queries and relevance annotations, we excluded semantically overlapping or hierarchically related event categories (\eg ``music'' and ``violin'').

Using these verified atomic sounds, we generate 10,000 composite clips by combining up to eight sound samples per clip. The corpus comprises 6,000 sequential compositions with non-overlapping events and 4,000 overlapping compositions with simultaneous events. We retain the event labels and their onset and offset times as composition metadata. These metadata are used to instantiate task-specific query templates (Figure~\ref{fig:fig1}) covering negation, order, overlap, duration, and combined constraints.

We determine query–audio relevance from the composition metadata and query constraints. GPT-5.5 provides an additional consistency check by assessing whether the event labels and timestamps satisfy the corresponding queries. This check operates on metadata and text rather than the synthesized audio. To complement this metadata-based validation, we conduct human listening evaluation in Section~\ref{main} to assess whether the query constraints can be reliably judged from the synthesized audio within the evaluated candidates.

\subsection{Natural Subtask Construction}
\label{natural_construction}


To evaluate generalization beyond synthetic audio, we construct the Natural subtask using 1,000 real-world audio clips sampled from AudioSet-Strong~\cite{hershey2021temporallystrong}. We manually write 100 reasoning-intensive queries following the combined-constraint design used in the Mix subtask. We use GPT-5.5 to assess query–audio relevance from sound-event annotations containing event labels and their onset and offset times. These annotations support evaluation of event presence, absence, order, overlap, and duration. Human annotators verify all positively labeled query–audio pairs by listening to the corresponding audio and checking each query constraint, including the absence of prohibited events. To check for missed positives, we randomly sample 50 queries and pool the top-10 clips retrieved by LCO-Embedding-7B and OmniEmbed-7B, removing duplicates within each query. Annotators listen to the pooled candidates initially labeled as negative and assess their relevance against the query constraints. No missed positives are found in this audited subset.


The resulting \ours benchmark comprises five synthetic subtasks, each with 200 queries over a shared corpus of 10,000 composite clips, and one natural subtask with 100 queries over 1,000 real-world clips. The average numbers of positives per query are 5.2, 1.6, 1.3, 46.4, 1.0, and 3.6 for Negation, Order, Overlap, Duration, Mix, and Natural, respectively.

\section{Experiment}
\label{main}
\subsection{Implementation Details}
We evaluate three types of \textaudio Retrieval approaches with 11 SOTA retrieval systems, together with a human baseline:




(1) \textbf{Two-Stage \textaudio Retrieval.} We evaluated a two-stage approach that sequentially applies audio-to-text models and text retrievers. Audio inputs are first converted into text descriptions using Qwen2-Audio~\cite{chu2024qwen2audiotechnicalreport} or Step-Audio~\cite{huang2025stepaudiounifiedunderstandinggeneration}, followed by text retrieval with BGE-M3~\cite{chen2025m3embeddingmultilingualitymultifunctionalitymultigranularity} or Qwen3-Embedding~\cite{zhang2025qwen3embeddingadvancingtext}.

(2) \textbf{CLIP-style Embedding Models.} We evaluated CLIP-style embedding models trained with contrastive learning, including CLAP~\cite{wu2024largescalecontrastivelanguageaudiopretraining}, AudioCLIP~\cite{guzhov2021audioclipextendingclipimage}, and Wav2CLIP~\cite{wu2022wav2cliplearningrobustaudio}.

(3) \textbf{MLLM-based Embedding Models.} We evaluated embedding models adapted from MLLMs, including e5-omni-7B~\cite{e5omni}, LCO-Embedding-3B/7B~\cite{xiao2025scalinglanguagecentricomnimodalrepresentation},  and OmniEmbed-7B~\cite{tevatron}.

(4) \textbf{Human Evaluation.} To assess subtask solvability and check for potential flaws in the synthetic data, we conduct a human evaluation with three annotators. Each annotator independently selects the most relevant clip from six candidates drawn from OmniEmbed-7B's top-ranked results, replacing one candidate with a ground-truth positive when none is present. This evaluation measures solvability within a restricted candidate pool; its results are therefore not directly comparable to other full-corpus retrieval baselines.




\subsection{Evaluation Metrics}
We report Accuracy@1 (Acc@1) and nDCG@10 to measure retrieval accuracy and ranking quality, respectively, with all scores scaled by 100. The overall score is the unweighted mean of both metrics across all six subtasks, summarizing performance across tasks and metrics. We omit Recall@1/5/10 due to space constraints, as it exhibits trends consistent with the reported metrics. Human evaluation uses Acc@1 to measure selection accuracy within the six-candidate pool.




\subsection{Main Results}

\noindent\textbf{Finding I --- Reasoning-intensive retrieval remains challenging for existing models despite being solvable for humans.} As shown in Table~\ref{tab:main_results}, all evaluated models perform poorly, while human evaluation confirms that the tasks are solvable. The best-performing model, OmniEmbed-7B, obtains an overall score of 20.7. Two-stage and CLIP-style models are largely ineffective. MLLM-based models perform comparatively better: in terms of Acc@1, OmniEmbed-7B achieves 24.5\% on Order and 28.0\% on Mix, while LCO-Embedding-7B reaches 29.5\% on Overlap and 14.5\% on Duration. Nevertheless, all models struggle substantially with Negation and Duration. This limitation is consistently observed in the Natural subtask, where the highest nDCG@10 is only 31.5\%, achieved by OmniEmbed-7B despite retrieving from a much smaller corpus. 


\par\smallskip
\noindent\textbf{Finding II --- Both captioning errors and reasoning limitations hinder two-stage methods.} 
The two-stage pipeline also performs poorly, achieving overall scores of only 1.9--3.6. Step-Audio performs better on the synthetic subtasks, whereas Qwen2-Audio performs better on Natural. Stronger text retrievers yield only modest gains, suggesting that errors in audio descriptions contribute to the poor retrieval performance. In an oracle analysis using Qwen3-Embedding, we replace each audio clip with a ground-truth textual description generated from its composition metadata. This raises Acc@1 to 65.0\%, 89.0\%, and 98.5\% on Order, Overlap, and Mix, respectively, while performance on Negation and Duration remains limited at 17.5\% and 38.0\%. Thus, imperfect event recognition explains part of the error, while failures to follow logical constraints persist without acoustic input.


\par\smallskip
\noindent\textbf{Finding III --- MLLM-based models perform better but remain far from satisfactory.}
MLLM-based models generally outperform other approaches, with OmniEmbed-7B achieving the highest overall score of 20.7. However, performance remains low, particularly on Negation and Duration. Scaling yields moderate gains: LCO-Embedding-7B improves over its 3B counterpart by 2.8 points in overall score. Manual inspection shows that top-ranked clips often contain the queried events but violate logical or temporal constraints.

\begin{figure}[t!]
\centering
\includegraphics[width=\columnwidth]{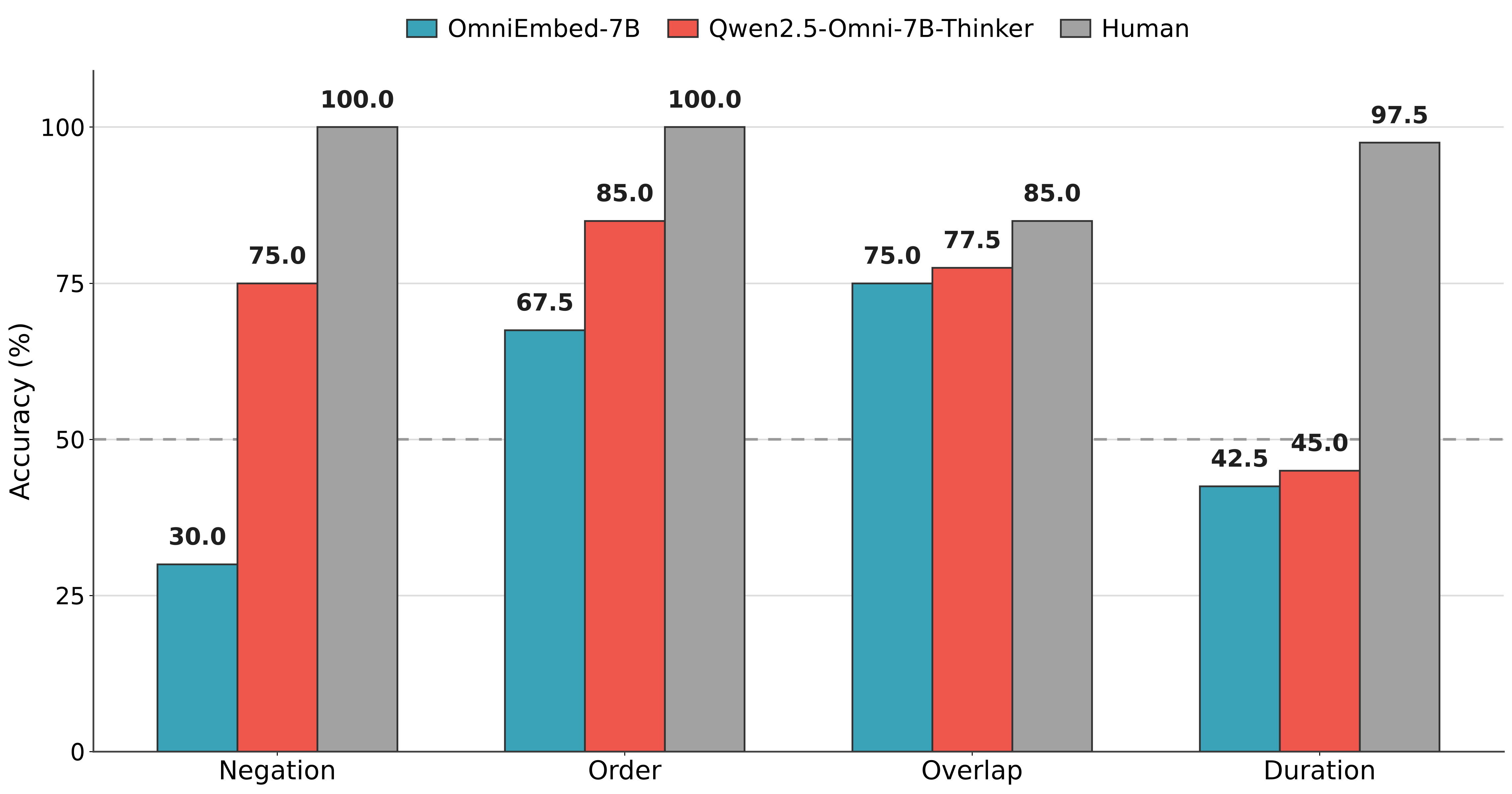}
\caption{Two-choice accuracy of OmniEmbed-7B, its Qwen2.5-Omni-7B-Thinker backbone, and human participants across the four core reasoning abilities. The dashed line indicates chance performance (50\%). Human and model performance are directly comparable in this setting, as both are evaluated using the same candidate sets.}
\label{fig:binary_reasoning}
\end{figure}





\section{Further Analysis}
To reduce the influence of sound-event matching, we conduct a controlled two-choice evaluation with 40 queries for each of the four core reasoning abilities. Both candidate clips contain the required sound events, but only one satisfies all query constraints. OmniEmbed-7B selects the candidate with the higher query–audio cosine similarity, while its generative backbone, Qwen2.5-Omni-7B-Thinker, and human annotators choose between the same two candidates. As shown in Figure 3, OmniEmbed-7B, the backbone, and humans achieve average accuracies of 53.8\%, 70.6\%, and 95.6\%, respectively. The backbone outperforms OmniEmbed-7B on Negation (75.0\% vs. 30.0\%) and Order (85.0\% vs. 67.5\%), while their scores on Overlap are similar (77.5\% vs. 75.0\%). On Duration, both models score below the 50\% chance level, whereas humans achieve 97.5\%.

These results suggest that generative reasoning capabilities are not consistently reflected in embedding-based retrieval performance, as illustrated by the gaps on Negation and Order. Both models struggle with Duration, indicating a shared difficulty in judging event duration. In contrast, humans achieve high accuracy across all four abilities, suggesting that the evaluated tasks are comparatively straightforward for human listeners.

\section{Conclusion}

In this paper, we present \ours, a benchmark designed to systematically evaluate reasoning-intensive text–audio retrieval, covering four core reasoning abilities across six synthetic and natural subtasks. Evaluation of 11 state-of-the-art retrieval systems reveals substantial gaps in these abilities. Controlled comparisons further reveal that OmniEmbed-7B underperforms its generative backbone on average in the two-choice evaluation, highlighting a gap between generative reasoning and embedding-based retrieval. We hope \ours will serve as a useful tool for advancing audio retrieval.

\section{Compliance with Ethical Standards}
The human evaluation involved only non-sensitive audio relevance judgments and collected no personal or identifiable information. No formal ethics review was sought for this minimal-risk study.


\bibliographystyle{IEEEbib}
\bibliography{custom}

\end{document}